\documentclass[10pt,twocolumn,letterpaper]{article}

\usepackage{cvpr}
\usepackage{times}
\usepackage{epsfig}
\usepackage{graphicx}
\usepackage{amsmath}
\usepackage{amssymb}
\usepackage{multirow}
\usepackage{subfig}

\usepackage[pagebackref=true,breaklinks=true,letterpaper=true,colorlinks,bookmarks=false]{hyperref}

\cvprfinalcopy 

\def\cvprPaperID{2799} 

\begin{document}

\title{CNN based Learning using Reflection and Retinex Models for Intrinsic Image Decomposition}

\author{Anil S. Baslamisli, Hoang-An Le, Theo Gevers\\
Informatics Institute\\
University of Amsterdam\\
{\tt\small \{a.s.baslamisli, h.a.le, th.gevers\}@uva.nl}
}

\maketitle

\begin{abstract} 
   Most of the traditional work on intrinsic image decomposition rely on deriving priors about scene characteristics. On the other hand, recent research use deep learning models as in-and-out black box and do not consider the well-established, traditional image formation process as the basis of their intrinsic learning process. As a consequence, although current deep learning approaches show superior performance when considering quantitative benchmark results, traditional approaches are still dominant in achieving high qualitative results. 
   
   In this paper, the aim is to exploit the best of the two worlds. A method is proposed that (1) is empowered by deep learning capabilities, (2) considers a physics-based reflection model to steer the learning process, and (3) exploits the traditional approach to obtain intrinsic images by exploiting reflectance and shading gradient information. The proposed model is fast to compute and allows for the integration of all intrinsic components. To train the new model, an object centered large-scale datasets with intrinsic ground-truth images are created.
   
   The evaluation results demonstrate that the new model outperforms existing methods. Visual inspection shows that the image formation loss function augments color reproduction and the use of gradient information produces sharper edges.
   
   Datasets, models and higher resolution images are available at \url{https://ivi.fnwi.uva.nl/cv/retinet}.
\end{abstract}

\section{Introduction}

Intrinsic image decomposition is the process of separating an image into its formation components such as reflectance (albedo) and shading (illumination)~\cite{Barrow}. Reflectance is the color of the object, invariant to camera viewpoint and illumination conditions, whereas
shading, dependent on camera viewpoint and object geometry, consists of different illumination effects, such as shadows, shading and inter-reflections. Using intrinsic images, instead of the original images, can be beneficial for many computer vision algorithms. For instance, for shape-from-shading algorithms, the shading images contain important visual cues to recover geometry, while for segmentation and detection algorithms, reflectance images can be beneficial as they are independent of confounding illumination effects.
Furthermore, intrinsic images are used in a wide range of computational photography applications, such as material recoloring~\cite{Ye,Meka0}, relighting~\cite{Beigpour,Duchene}, retexturing~\cite{Bousseau,Yan} and stylization~\cite{Ye}.

Most of the pioneering work on intrinsic image decomposition, such as \cite{Barron0,Barrow,tappen,Weiss}, rely on deriving priors about scene characteristics to understand the physical interactions of objects and lighting in a scene. In general, an optimization approach is taken imposing constraints on reflectance and shading intrinsics for a pixel-wise decomposition. \cite{Land} introduces the well-known Retinex algorithm which is based on the assumption that larger gradients in an image usually correspond to reflectance changes, whereas smaller gradients are more likely to correspond to illumination changes. In addition to the traditional work, more recent research focuses on using deep learning (e.g. CNN) models \cite{narihia1,shapenet}. However, these deep learning-based methods do not consider the well-established, traditional image formation process as the basis of their intrinsic learning process. Deep learning is used as in-and-out black box, which may lead to inadequate or restricted results. Furthermore, the contribution and physical interpretation of what the network learned is often difficult to interpret. As a consequence, although current deep learning approaches show superior performance when considering quantitative benchmark results, traditional approaches are still dominant in achieving high qualitative results. Therefore, the goal of this paper is to exploit the best of the two worlds. A method is proposed that (1) is empowered by deep learning capabilities, (2) considers a physics-based {\it reflection model} to steer the learning process, and (3) exploits the traditional approach to obtain intrinsic images by exploiting reflectance and shading {\it gradient} information.

To this end, a physics-based convolutional neural network, \textit{IntrinsicNet}, is proposed first. A standard CNN architecture is chosen to exploit the dichromatic reflection model~\cite{Shafer} as a standard reflection model to steer the training process by introducing a physics-based loss function called the \textit{image formation loss}, which takes into  account the reconstructed image of the predicted reflectance and shading images. The goal is to analyze the contribution of exploiting the image formation process as a constraining factor in a standard CNN architecture for intrinsic image decomposition. Then, we propose the \textit{RetiNet}, which is a two-stage Retinex-inspired convolutional neural network which first learns to decompose (color) image gradients into intrinsic image gradients i.e. reflectance and shading gradients. Then, these intrinsic gradients are used to learn the CNN to decompose, at the pixel, the full image into its corresponding reflectance and shading images. 

The availability of annotated large-scale datasets is key to the success of supervised deep learning methods. However, the largest publicly available dataset with intrinsic image ground-truth has around a thousand of redundant images taken from an animated cartoon-like short film \cite{sintel}. Therefore, to train our CNN's, we introduce a large-scale dataset with intrinsic ground-truth images: a synthetic dataset with man-made objects. The dataset consists of around 20,000 images. Rendered with different environment maps and viewpoints, the dataset provides a variety of possible images in indoor and outdoor scenes.

In summary, our contributions are: (1) a standard CNN architecture \textit{IntrinsicNet} incorporating the \textit{image formation loss} derived by a physics-based reflection model, (2) a new two-stage Retinex-inspired convolutional neural network \textit{RetiNet} exploiting \textit{intrinsic gradients} for image decomposition at the pixel, (3) gradient (re)integration (inverse problem) where images are integrated based on intrinsic gradients by a set of simple convolutions rather than complex computations (e.g. Poisson), and (4) a large-scale synthetic object-centered dataset with intrinsic ground-truth images.

\section{Related Work}

Since there are multiple unknowns and multiple solutions to recover the pixel intrinsics, intrinsic image decomposition is an ill-posed and under-constrained problem \cite{Gehler,Shen2}. Therefore, most of the related work derive priors about the scene characteristics and impose constraints on the reflectance and shading maps. Usually an optimization procedure is used enforcing imaging constraints for pixel-wise decomposition.
One of the earliest and most successful methods is the Retinex algorithm~\cite{Land}. This approach considers that the reflectance image is piece-wise constant and that the shading image varies smoothly. The algorithm assumes that larger derivatives
in an image correspond to reflectance changes, and that the smaller ones correspond to illumination changes. This approach is extended to color images~\cite{Funt} by exploiting the chromaticity information, which is invariant to shading cues. Since then, most of the (traditional) related work continued to focus on understanding the physical interactions, geometries of the objects, and lighting cues by inferring priors. Priors that are used to constrain the inference problem are
based on texture cues~\cite{Shen,Zhao}, sparsity of reflectance~\cite{Gehler,Shen2}, user in the loop~\cite{Bousseau,Shen3},
and depth cues~\cite{Barron0,Chen,Lee}. Other methods use multiple images~\cite{Laffont1,Matsushita,Weiss}, where reflectance is considered as the constant factor
and illumination the changing one. These methods produce promising results as they disambiguate the decomposition. However, their applicability is limited by the use of priors. 

\textbf{Supervised Deep Learning:} 
Deep convolutional neural networks are very successful for various computer vision tasks, such as image classification~\cite{vggnet} 
and object detection~\cite{rcnn}. However, the success of supervised deep learning is dependent on the availability of annotated large-scale datasets~\cite{coco,imagenet}. Collecting and annotating large-scale datasets takes considerable time and effort for most of the deep learning related classification tasks. However, these images are mostly collected from the internet, which makes them easily accessible by nature. On the other hand, the process of data generation and annotation is more difficult for dense (pixel-wise) prediction tasks, such as semantic segmentation, optical flow estimation and intrinsic image decomposition. For those tasks, generating and annotating synthetic data
in an automated fashion is relatively easier than collecting and annotating real world data. The use of synthetic data has proven to produce competitive performance~\cite{mayer,synthia}. For real data, collecting and generating ground-truth intrinsic images is only possible in a fully-controlled laboratory settings. It requires a delicate setting to separate intrinsic images step by step from the original image. This process requires excessive effort and time ~\cite{mit}. At this moment, the only existing dataset with real world images and corresponding ground-truth intrinsics contains as few as 20 object-centered images ~\cite{mit}. As a result, intrinsic image research using supervised deep learning is dependent on synthetic datasets. ~\cite{sintel} provides a scene-level 3D animated
cartoon-like short film with intrinsic image ground-truths. Although the dataset has around a thousand images, it is demonstrated by~\cite{kim,narihia1} that this dataset of synthetic ground-truth is useful for training convolutional neural networks. Further, ~\cite{shapenet} provides a large dataset of non-Lambertian synthetic objects (around 3 million images). Using this dataset, ~\cite{shapenet} achieves state of the art results by training an encoder-decoder based convolutional neural network. However, their dataset is not publicly available, yet. ~\cite{iiw} provides relative reflectance comparisons over point pairs of real world indoor scenes. These indoor scenes are annotated by crowd-sourcing. Although it does not have ground-truth intrinsic images, it is effective in learning priors and relationships in a data-driven manner~\cite{narihia0,zhou,zoran}.

Supervised deep learning, trained on large scale datasets, achieves state of the art results on different benchmarks. However, they ignore physics-based characteristics of  the intrinsic image formation process. These methods use deep learning as a black box. The use of reflection models are used by traditional methods. However, traditional methods do not exploit the learning power of CNNs. ~\cite{narihia1} argues that the learning model should consider both patch level information and the overall gist of the scene. In more recent work, the proposed model is based on the assumption that the intrinsic components are highly correlated ~\cite{shapenet}. Their training data is generated in a physics-based manner as well, including a specular component, but they do not explicitly embed a physics-based image formation loss. Another recent work \cite{kim} uses an image formation component in their unary term for CRF (for the optimization process, not in the learning process itself), but their training data (Sintel) was not created in a physics-based manner. Nonetheless, none of proposed deep learning methods consider the image formation process for consistent decomposition during training, nor a Retinex driven gradient separation approach~\cite{bell,Gehler,mit,Shen,tappenn,tappen}. Retinex has a solid background in intrinsic image decomposition. Therefore, this paper combines the best of the two worlds: supervised deep learning based on reflection and Retinex models.

\begin{figure*}[t]
\includegraphics[scale=0.32]{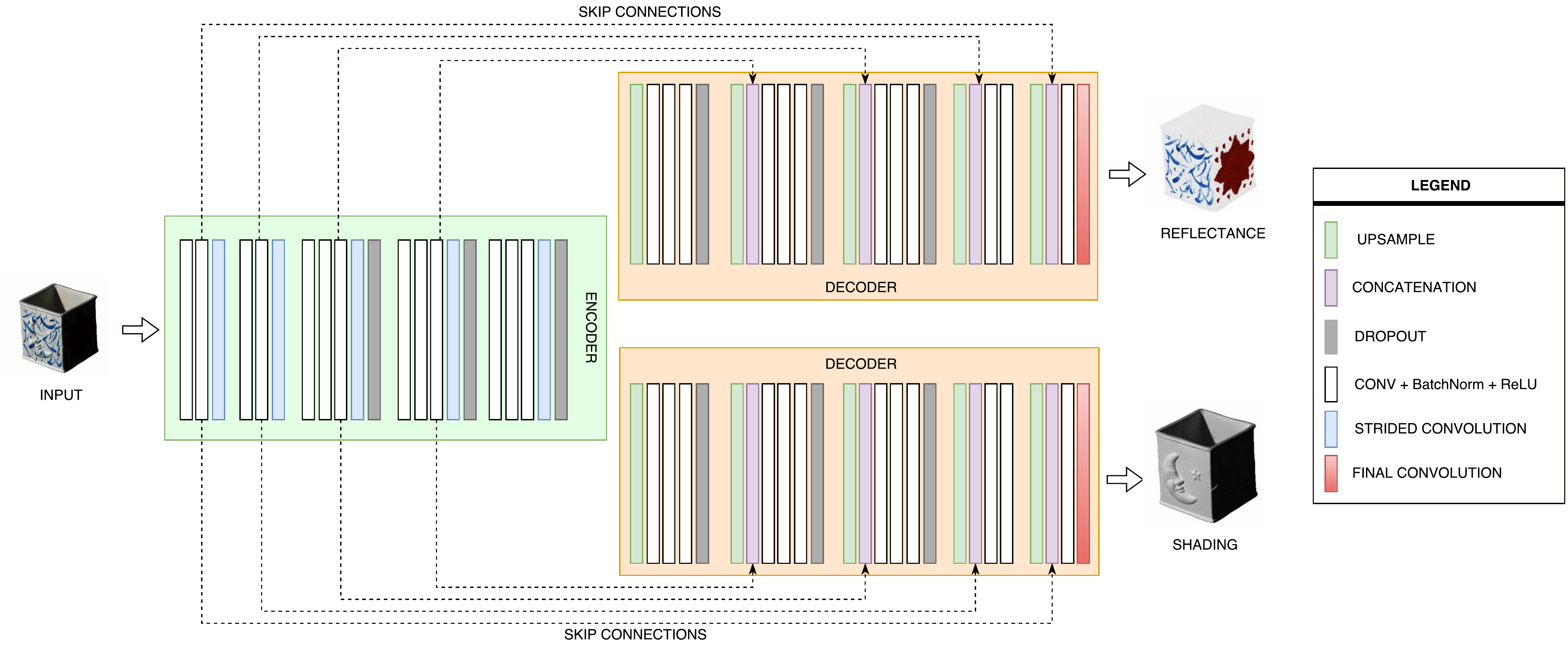}
\centering
\caption{IntrinsicNet model architecture with one shared encoder and two separate decoders: one for shading and one for reflectance prediction.
Encoder part contains both shading and reflectance characteristics. The decoder parts aim to disentangle those features.}
\label{fig:shared}
\end{figure*}

\section{Approach}
In this section, the image formation model is described first. Then, we propose an encoder-decoder CNN, called IntrinsicNet, which is a convolutional neural network based on the reflection model by introducing the image formation loss. Finally, we propose a new CNN architecture, RetiNet, which is a Retinex-inspired scheme that exploits image gradients in combination with the image formation loss.

\subsection{Image Formation Model}
The dichromatic reflection model~\cite{Shafer} describes a surface as a composition of the body $I_b$ (diffuse) and specular $I_s$ (interface) reflectance:
\begin{equation} \label{eq:brdf}
I = I_b + I_s.
\end{equation}
Then, the pixel value, measured over the visible spectrum $\omega$, is expressed by:
\begin{equation} \label{eq:brdf2}
\begin{aligned}
I = m_b(\vec{n}, \vec{s}) \int_{\omega}^{} f_{c}(\lambda)\;e(\lambda)\;\rho_{b}(\lambda)\; \mathrm{d}\lambda \;+ \\
m_s(\vec{n}, \vec{s}, \vec{v}) \int_{\omega}^{} f_{c}(\lambda)\;e(\lambda)\;\rho_{s}(\lambda)\; \mathrm{d}\lambda,
\end{aligned}
\end{equation}
where $\vec{n}$ is the surface normal, $\vec{s}$ is the light source direction, and $\vec{v}$ is the viewing/camera direction. $m$ is a function of the geometric dependencies. Furthermore, $\lambda$ is the wavelength, $f_{c}(\lambda)$ is
the camera spectral sensitivity, $e(\lambda)$ defines the spectral power distribution of the illuminant, $\rho_{b}$ characterizes the diffuse surface reflectance, and $\rho_{s}$ is the specular reflectance with Fresnel reflection.

Assuming a linear sensor response and narrow band filters ($\lambda_I$), Equation (\ref{eq:brdf2}) is as follows:
\begin{equation} \label{eq:nb}
\begin{aligned}
I = m_b(\vec{n}, \vec{s})\;e(\lambda_I)\;\rho_{b}(\lambda_I)\;+\\
m_s(\vec{n}, \vec{s}, \vec{v})\;e(\lambda_I)\;\rho_{s}(\lambda_I).
\end{aligned}
\end{equation}
\noindent Then, under the assumption of body (diffuse) reflection, the decomposition of the observed image $I(\vec{x})$ at position $\vec{x}$ can be approximated as the element-wise
product of its reflectance $R(\vec{x})$ and shading $S(\vec{x})$ intrinsics:
\begin{equation} \label{eq:iid}
I(\vec{x}) = R(\vec{x}) \times S(\vec{x}),
\end{equation}
where $S(\vec{x})$ can be Lambertian i.e. the dot product of $\vec{n}$ and $\vec{s}$ at location $\vec{x}$. In Equation (\ref{eq:nb}), $e(\lambda_I)$ is modeled as a single, canonical light source. We can extend the model for a non-canonical light source as follows:
\begin{equation} \label{eq:iid2}
I(\vec{x}) = R(\vec{x}) \times S(\vec{x}) \times E(\vec{x}),
\end{equation}
where $E(\vec{x})$ describes the color of the light source at position $\vec{x}$. The model for a global, non-canonical light source is described by:
\begin{equation} \label{eq:iid3}
I(\vec{x}) = R(\vec{x}) \times S(\vec{x}) \times E.
\end{equation}
\noindent Equation (\ref{eq:iid}) is extended to non-diffuse reflection by adding the specular (surface) term $H(\vec{x})$:
\begin{equation} \label{eq:iid4}
I(\vec{x}) = R(\vec{x}) \times S(\vec{x}) + H(\vec{x}).
\end{equation}
and for a non-canonical light source by:
\begin{equation} \label{eq:iid5}
I(\vec{x}) = R(\vec{x}) \times S(\vec{x}) \times E(\vec{x}) + H(\vec{x}) \times E(\vec{x}).
\end{equation}
Finally, for a global, non-canonical light source we obtain:
\begin{equation} \label{eq:iid6}
I(\vec{x}) = R(\vec{x}) \times S(\vec{x}) \times E + H(\vec{x}) \times E.
\end{equation}
In the next section, the reflection model is considered to introduce different image formation losses within an encoder-decoder CNN model for intrinsic image decomposition.

\subsection{IntrinsicNet: CNN driven by Reflection Models} 

In this section, a physics-based deep learning network, IntrinsicNet, is proposed. We use a standard CNN architecture to constrain the training process by introducing a physics-based loss. The reason of using a standard CNN architecture is to analyze whether it is beneficial to constrain the CNN by the reflection model. Therefore, an end-to-end trainable encoder-decoder CNN is considered. These type of CNNs yield good results in most of the pixel-wise dense prediction tasks \cite{shapenet,demon}. An architecture is adopted with one shared encoder and two separate decoders: one for shading prediction and one for reflectance prediction. The features learned by the encoder stage contain both shading and reflectance cues. The purpose of the decoder parts is to disentangle those features.
Figure \ref{fig:shared} illustrates our model. Obviously, the architecture can be extended by considering more image formation factors (e.g. the light source or highlights) by adding the corresponding decoder blocks.

To train the model, we use the standard $L_2$ reconstruction loss. Let $\hat{J}$ be the ground-truth intrinsic image and $J$ be the prediction of the network. Then, the reconstruction loss $\mathcal{L}_{RL}$ is given by:
\begin{equation} \label{eq:L2}
\mathcal{L}_{RL}(J, \hat{J}) = \frac{1}{n} \sum_{\vec{x},c}^{} ||\hat{J} - J||^{2}_{2},
\end{equation}
where $\vec{x}$ denotes the image pixel, $c$ the channel index and $n$ is the total number of evaluated pixels. In our case, the final, combined loss $\mathcal{L}_{CL}$ is composed of 2 distinct loss functions, one for reflectance reconstruction
$\mathcal{L}_{RL_R}$ and one for shading reconstruction $\mathcal{L}_{RL_S}$:
\begin{equation} \label{eq:L2_shared}
\begin{aligned}
\mathcal{L}_{CL}(R, \hat{R}, S, \hat{S}) = \gamma_R \; \mathcal{L}_{RL_R}(R, \hat{R}) \;+ \\ \gamma_S \; \mathcal{L}_{RL_S}(S, \hat{S}),
\end{aligned}
\end{equation}
where the $\gamma$s are the corresponding weights. In general, this type of network may generate color artifacts and blurry reflectance maps \cite{shapenet,demon}. The goal of the image formation loss is to increase the color reproduction quality because of the physics constraint. 

More precisely, the image formation loss $\mathcal{L}_{IMF}$ takes into account the reconstructed image of the predicted reflectance and shading images. That is in addition to the $RGB$ input image. Hence, this loss imposes the reflection model constraint of Equation (\ref{eq:iid}):
\begin{equation} \label{eq:L2_physics}
\mathcal{L}_{IMF}(R, S, I) = \gamma_{IMF} \; \mathcal{L}_{RL_{IMF}}((R \times S), \;I)
\end{equation}
where $I$ is the input image. Thus, the final loss of the IntrinsicNet becomes:
\begin{equation} \label{eq:L2_final}
\begin{aligned}
\mathcal{L}_{FL}(I, R, \hat{R}, S, \hat{S}) = \mathcal{L}_{CL}(R, \hat{R}, S, \hat{S}) \;+ \\ \mathcal{L}_{IMF}(R, S, I).
\end{aligned}
\end{equation}
Note that the image formation loss is not limited to Equation (\ref{eq:iid}). Any intrinsic image Equation (\ref{eq:iid}-\ref{eq:iid6}) can be used depending on the intrinsic problem at hand. For example, the loss function for the full reflection model $\mathcal{L}_{FRM}$ is as follows:
\begin{equation} \label{eq:fullmodel}
\begin{split}
\mathcal{L}_{FRM}(*) = \gamma_R \; \mathcal{L}_{RL_R}(R, \hat{R}) \;+ \gamma_S \; \mathcal{L}_{RL_S}(S, \hat{S}) \; +\\ \gamma_H \; \mathcal{L}_{RL_H}(H, \hat{H}) \;+ \gamma_E \; \mathcal{L}_{RL_E}(E, \hat{E}) \;+ \\ \gamma_{IMF} \; \mathcal{L}_{RL_{IMF}}((R \times S \times E + H \times E), \;I).
\end{split}
\end{equation}

The image formation loss function is designed to augment the color reproduction. To augment both {\em color reproduction} and {\em edge sharpness}, in the next section, a two-stage Retinex-inspired CNN architecture is described which uses intrinsic gradients (for edge sharpness) and the image formation loss (for color reproduction).

\subsection{RetiNet}

\begin{figure*}[t]
\includegraphics[scale=0.42]{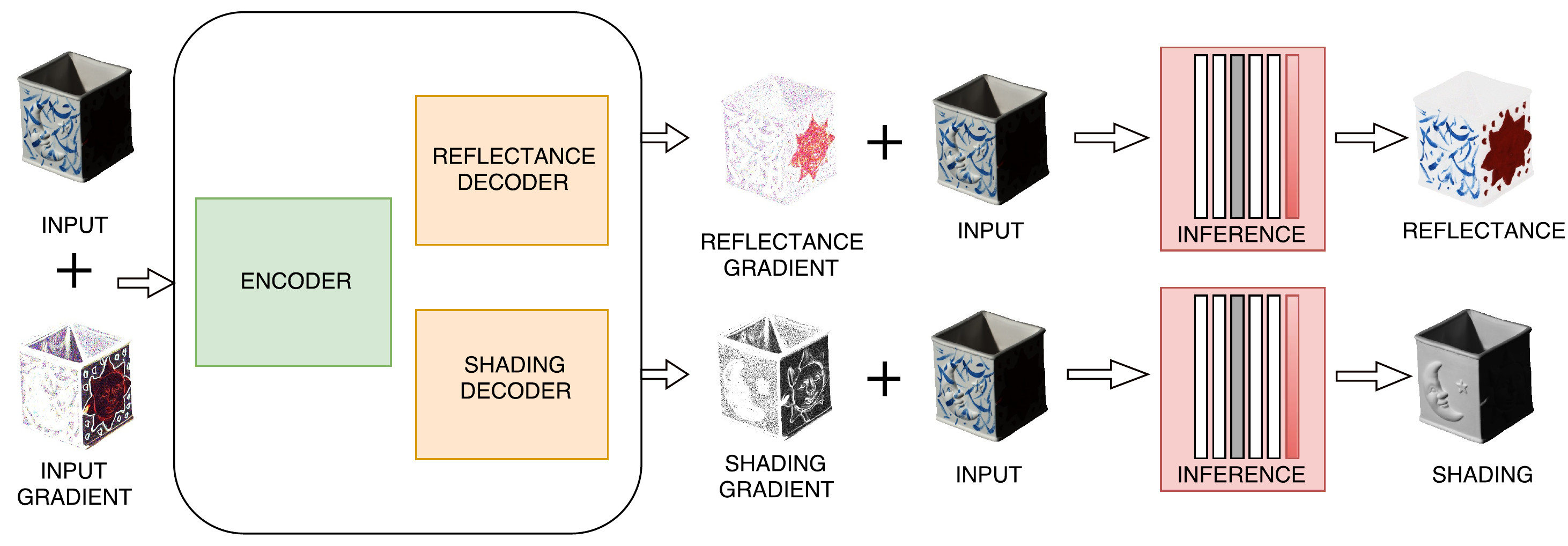}
\centering
\caption{RetiNet model architecture. Refer to Figure \ref{fig:shared} for layer types and encoder-decoder sub-network details. Instead of generating intrinsic image pixel values, the encoder-decoder network is trained to separate (color) image gradients into intrinsic image gradients. Then, for gradient re-integration part, the input image is concatenated with predicted intrinsic gradients and forwarded to a fully convolutional sub-network to perform the actual pixel-wise intrinsic image decomposition.}
\label{fig:final}
\end{figure*}

In this section, we exploit how a well-established, traditional approach such as Retinex can be used to steer the design of a CNN architecture for intrinsic image decomposition. Therefore, we propose the RetiNet model. In fact, the RetiNet architecture is a two-stage Retinex-inspired CNN that exploits gradient information in combination with the image formation loss. Actually, most of the traditional approaches follow the successful Retinex findings of using gradient separation~\cite{bell,Gehler,mit,Shen,tappenn,tappen}. In contrast to threshold-driven gradient separation, the goal of our network is to learn intrinsic gradients directly from data avoiding hard-coded thresholds. Further, for the re-integration process, we propose a series of simple convolutions to efficiently compute the intrinsic images separately. That is in contrast to other methods which try to find, by complex computations, the pseudo-inverse of an unconstrained system of derivatives, or to solve the Poisson equation. 

Image gradients are calculated by taking the intermediate difference between neighboring pixels; horizontal ($G_x$) and vertical ($G_y$) separately. Finally, the gradient magnitude ($G$) is given as the square root of the sum of squares of the horizontal and the vertical components of the gradient: 
\begin{equation}
G =\sqrt {{G_x}^2+{G_y}^2}
\end{equation}
This operation is carried out for each color channel individually. Then, the input is formed by concatenating the $RGB$ image with its gradients per color channel, resulting in a 6 channel input.
In this way, the network is assisted by image gradients. Finally, the encoder-decoder network is trained to separate color image gradients to intrinsic image gradients by using Equation (\ref{eq:L2_shared}): 
\begin{equation} \label{eq:1stphase}
\mathcal{L}_{S1} = \mathcal{L}_{CAL}(\nabla R, \nabla \hat{R}, \nabla S, \nabla \hat{S}),
\end{equation}
where $\nabla$ denotes the image gradient. For the first stage, we use the IntrinsicNet architecture described in the previous section. For the second stage, the input image is concatenated with the predicted intrinsic gradients this time. The newly formed input is provided to a fully convolutional sub-network to perform the actual decomposition by using Equation (\ref{eq:L2_final}) with the intrinsic loss. Figure \ref{fig:final} illustrates our RetiNet model. 

\cite{genericmodel} has some similarities with our gradient-based model. However, our model differs from theirs in several ways. Instead of relying on edge information, we directly use gradient information. Further, we aim to learn to separate gradients into different components, whereas their method only has one component. Finally, we propose a series of simple convolutions for the reintegration part, while they use an encoder-decoder based network with deconvolutions.

\section{Experiments}

\subsection{New Synthetic Dataset of Man-made Objects}

For our experiments, large scale datasets are needed to train the networks. Unfortunately, the intrinsic, synthetic ground-truth dataset of \cite{shapenet} is not publicly available (yet). For comparison reasons, we created a new dataset following the one described by \cite{shapenet}. We randomly sample around 20,000 3D models obtained from the ShapeNet dataset \cite{shapenet_dataset} for training. To create more variation and to decouple the correlation between image shape and texture, the texture of each component in a model is replaced by a random color. To enforce the lighting model, we apply a diffuse bidirectional scattering distribution function (BSDF) on the object surface with a random roughness parameter. The rendering is performed by the physics-based Blender Cycles\footnote{\url{https://www.blender.org/}}. Different environment maps are used to render the models at random viewpoints sampled from the upper hemisphere as conducted in \cite{shapenet}. To guarantee the relationship between reflectance and shading, the Cycles pipeline is modified to obtain the output image, its corresponding reflectance, and the shading map in high-dynamic range without gamma-correction. Since the images are taken from objects, the final dataset of around 20,000 images are object-centered. The object-centered dataset represent man-made objects. 

\begin{figure}[t]
\includegraphics[scale=0.45]{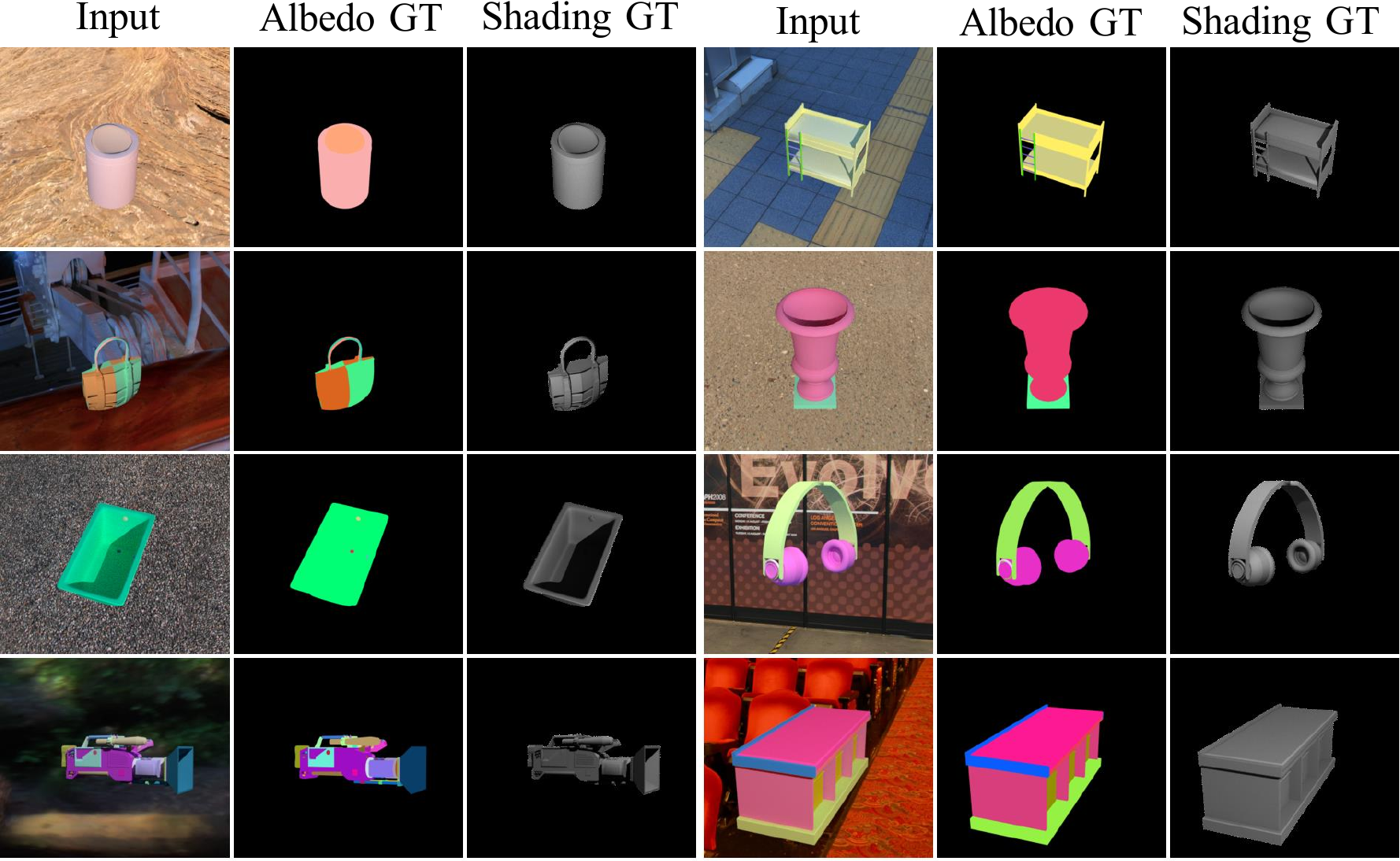}
\centering
\caption{Overview of the synthetic dataset. Different environment maps are used to render the models for realistic appearance. }
\label{fig:dataset}
\end{figure}

An overview of the datasets is given in Figure \ref{fig:dataset}. Rendered with different environment maps and viewpoints, the dataset provides a variety of possible images in indoor and outdoor scenes.

\subsection{Error Metrics}
To evaluate and compare our approach, metrics are chosen which are commonly used in the field. First, the results are evaluated in terms of the mean squared error (MSE) between the ground-truth intrinsic images and the measured ones. Following \cite{mit}, absolute brightness of each image is adjusted to minimize the error. Further, the local mean squared error (LMSE) \cite{mit} is chosen which is computed by aggregating the MSE scores over all local regions of size $k \times k$ with steps of $k/2$. Following the setup of \cite{mit}, all the results in the paper use $k$ = 20. The LMSE scores of the intrinsic images are averaged and normalized to make the maximum possible error equal to 1. To evaluate the perceptual visual quality of the results, the dissimilarity version of the structural similarity index (DSSIM) \cite{Chen} is taken.

\subsection{Implementation Details}
For the encoder network, the VGG16 architecture~\cite{vggnet} is used by removing the fully-connected layers. Moreover, for dimensionality reduction, the max-pooling layers are replaced by convolutional layers with stride 2. In this way, our model learns its customized spatial down-sampling and is fully convolutional. For the decoder network,   the encoder part is mirrored. The strided convolutional layers are inverted by a 4$\times$4 deconvolution with stride 2. Furthermore, we follow~\cite{skipconnection} and use skip-layer connections to pass image details to the top layers. The connections are linked between the convolutional layers before down-sampling of encoder blocks, and the corresponding deconvolutional layers of the decoder part, except between the last block of the encoder and the first block of the decoder. Moreover, batch normalization~\cite{batchnorm} is applied after each convolutional layer, except for the last layer of the decoders and the inference net of RetiNet (prediction results) to speed up the convergence and to maintain the gradient flow. The inference net has convolution kernels of 3$\times$3 and the layers have [64, 128, 128, 64] feature maps, respectively. Our models are implemented using the stochastic gradient descent optimizer with learning rate of 1e-5 and momentum of 0.9. A polynomial decay is applied to the learning rate to reach a final learning rate of 1e-7. Convolution weights are initialized by using \cite{msra} with a weight decay of 0.0005, whereas deconvolution weights are initialized randomly from a normal distribution with mean of 0 and standard deviation of 1. Moreover, the input size is fixed to 120$\times$160 pixels and the batch size is fixed at 16 for all experiments. Throughout all experiments, we randomly flip, vertical or horizontal, and shift images by a random factor of [-20, 20] pixels horizontally and vertically to generate additional training samples (data augmentation).

\section{Evaluation}


\begin{figure}[t]
\includegraphics[scale=0.7]{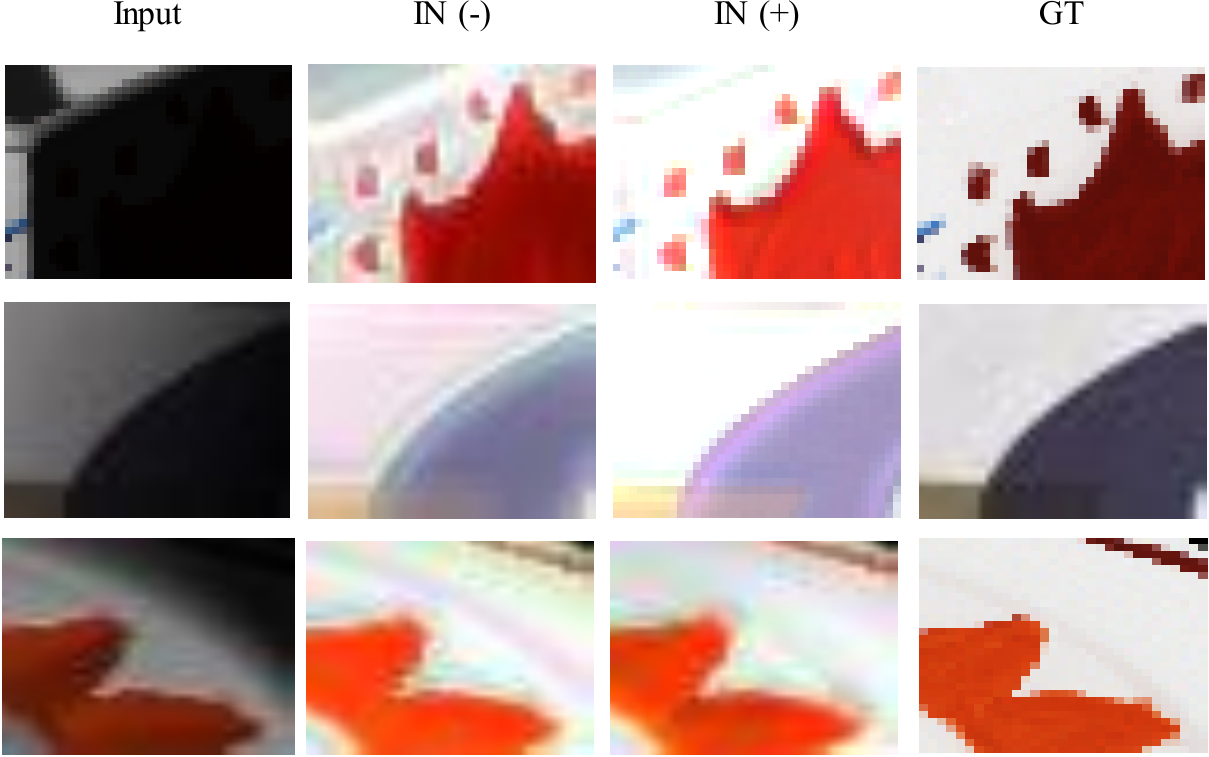}
\centering
\caption{MIT intrinsic benchmark differentiated by the use of the image formation loss. IN(+/-) denotes the IntrinsicNet with/without the image formation loss. The image formation loss suppresses color artifacts and halo effects.}
\label{fig:patchmit}
\end{figure}

\subsection{Image Formation Loss}
Figure \ref{fig:patchmit} shows detailed views of a patch, demonstrating the benefits of the image formation loss. It can be derived that the image formation loss suppresses color artifacts and halo effects. Furthermore, Table \ref{tab:imageFormationLoss} shows the quantitative evaluation results of our IntrinsicNet with and without the image formation loss ($\mathcal{L}_{IMF}$). The experiments on the MIT intrinsic benchmark show that the image formation loss constrains the model to obtain improved color reproduction as expressed quantitatively by the DSSIM metric. In addition, the model with the image formation loss obtains better results for the MSE and LMSE metrics on average. On the ShapeNet test set, the model with the image formation loss achieves similar performance for MSE and LMSE. On DSSIM, it produces proper results for albedo prediction. Considering the generalization ability and the effect on a unseen real-world dataset, it can be observed that the network with image formation loss achieves best performance for all metrics. It shows the positive contribution of exploiting the image formation process as a constraining factor in a standard CNN approach for intrinsic image decomposition.

\begin{table}[h]
  \centering
  \scalebox{0.65}{
  \begin{tabular}{ c||c|c||c|c||c|c }
    \multirow{2}{*}{ } &
      \multicolumn{2}{c||}{MSE}  &
      \multicolumn{2}{c||}{LMSE} &
      \multicolumn{2}{c}{DSSIM} \\
     & Albedo & Shading & Albedo & Shading & Albedo & Shading \\ \hline
    $^{*}$Without $\mathcal{L}_{IMF}$ & \textbf{0.0045} & 0.0062 & 0.0309 & 0.0326 & 0.0940 & 0.0704\\ \hline
    $^{*}$With $\mathcal{L}_{IMF}$ & 0.0051 & \textbf{0.0029} & \textbf{0.0295} & \textbf{0.0157} & \textbf{0.0926} & \textbf{0.0441} \\ \hline \hline \hline
    $^{+}$Without $\mathcal{L}_{IMF}$ & \textbf{0.0005} & \textbf{0.0007}  & 0.0300 & \textbf{0.0498} & 0.0075 & \textbf{0.0082} \\ \hline
    $^{+}$With $\mathcal{L}_{IMF}$ & \textbf{0.0005} & \textbf{0.0007} & \textbf{0.0297} & 0.0505 & \textbf{0.0072} & 0.0084 \\
  \end{tabular}}
  \newline
  \caption {Evaluation results of the IntrinsicNet with and without image formation loss on the MIT intrinsic benchmark ($*$) and the ShapeNet test set ($+$). The image formation loss constrains the model to obtain better DSSIM performance. At the same time, it outperforms other models considering the MSE and LMSE metrics on real world images.}
  \label{tab:imageFormationLoss}
\end{table}

\subsection{ShapeNet Dataset}
\begin{figure}[t]
\includegraphics[scale=0.92]{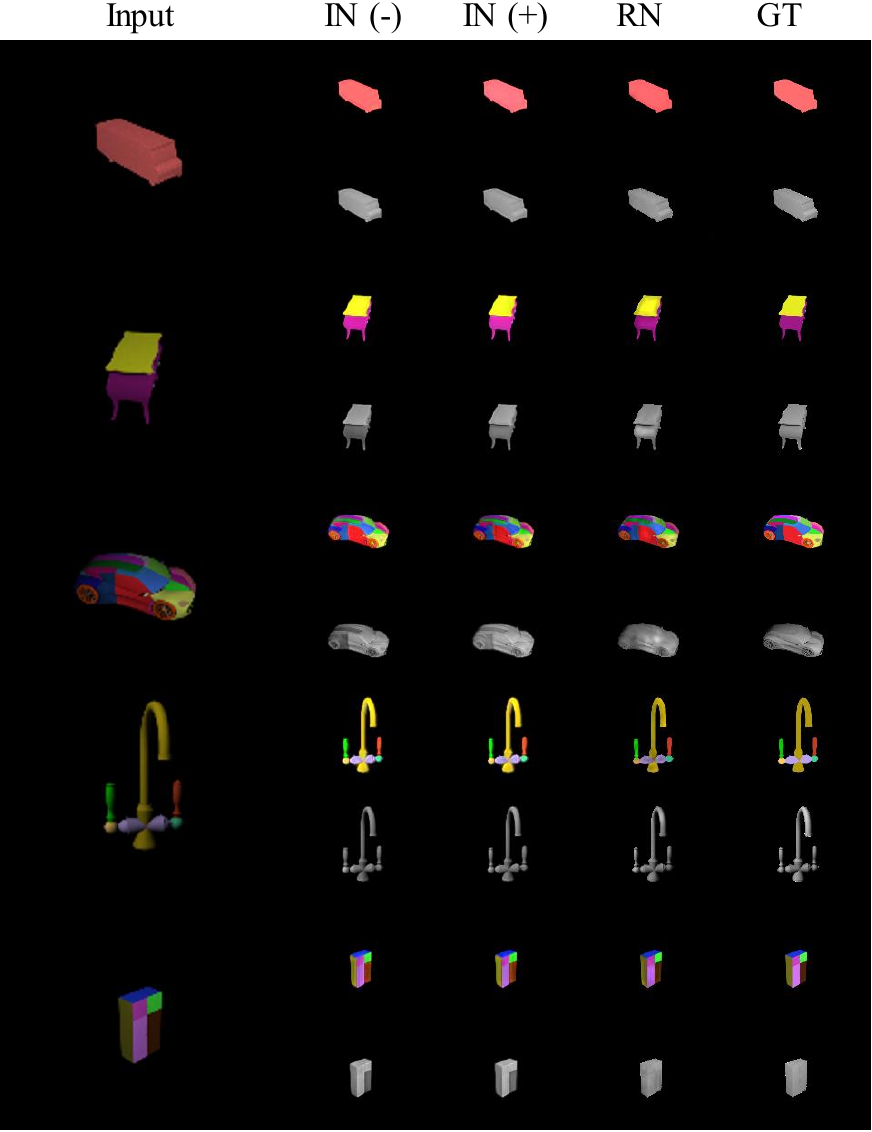}
\centering
\caption{Evaluation results on the synthetic test set. All proposed models produce decent intrinsic image compositions. IN(+/-) denotes the IntrinsicNet with/without the image formation loss, and RN denotes the RetiNet model.}
\label{fig:shapenetresults}
\end{figure}

We now test our models on the ShapeNet test partition. We follow the approach of \cite{shapenet} and randomly pick 1 image per test model, resulting in 7000 test images. For all experiments, the same test set is  used. Table \ref{tab:shapenetresults} shows the quantitative evaluation results of the synthetic test set of man-made objects. Figure \ref{fig:shapenetresults} displays (visual) comparison results. Our proposed methods yield better results on the test set. Moreover, our RetiNet model outperforms all by a large margin. Visual comparison results show that all of our proposed models are capable of producing decent intrinsic image compositions on the test set. 



\begin{table}[h]
  \centering
  \scalebox{0.65}{
  \begin{tabular}{ c||c|c||c|c||c|c }
    \multirow{2}{*}{ } &
      \multicolumn{2}{c||}{MSE}  &
      \multicolumn{2}{c||}{LMSE} &
      \multicolumn{2}{c}{DSSIM} \\
       & Albedo & Shading & Albedo & Shading & Albedo & Shading \\ \hline
       DirectIntrinsics\cite{narihia1} & 0.1487 & 0.0505 & 0.6868 & 0.3386 & 0.0475 & 0.0361 \\
    ShapeNet\cite{shapenet} & 0.0023 & 0.0037 & 0.0349 & 0.0608 & 0.0186 & 0.0171 \\ \hline \hline
    IntrinsicNet & 0.0005 & 0.0007 & 0.0297 & 0.0505 & 0.0072 & 0.0084 \\ 

    RetiNet & \textbf{0.0003} & \textbf{0.0004} & \textbf{0.0205} & \textbf{0.0253} & \textbf{0.0052} & \textbf{0.0064} \\  
  \end{tabular}}
  \newline
  \caption {Evaluation results on ShapeNet. Our proposed methods yield better results on the test set. Moreover, our RetiNet model outperforms all by a large margin.}
  \label{tab:shapenetresults}
\end{table}

\subsection{MIT Intrinsic Dataset}
\begin{figure}[t]
\includegraphics[scale=0.48]{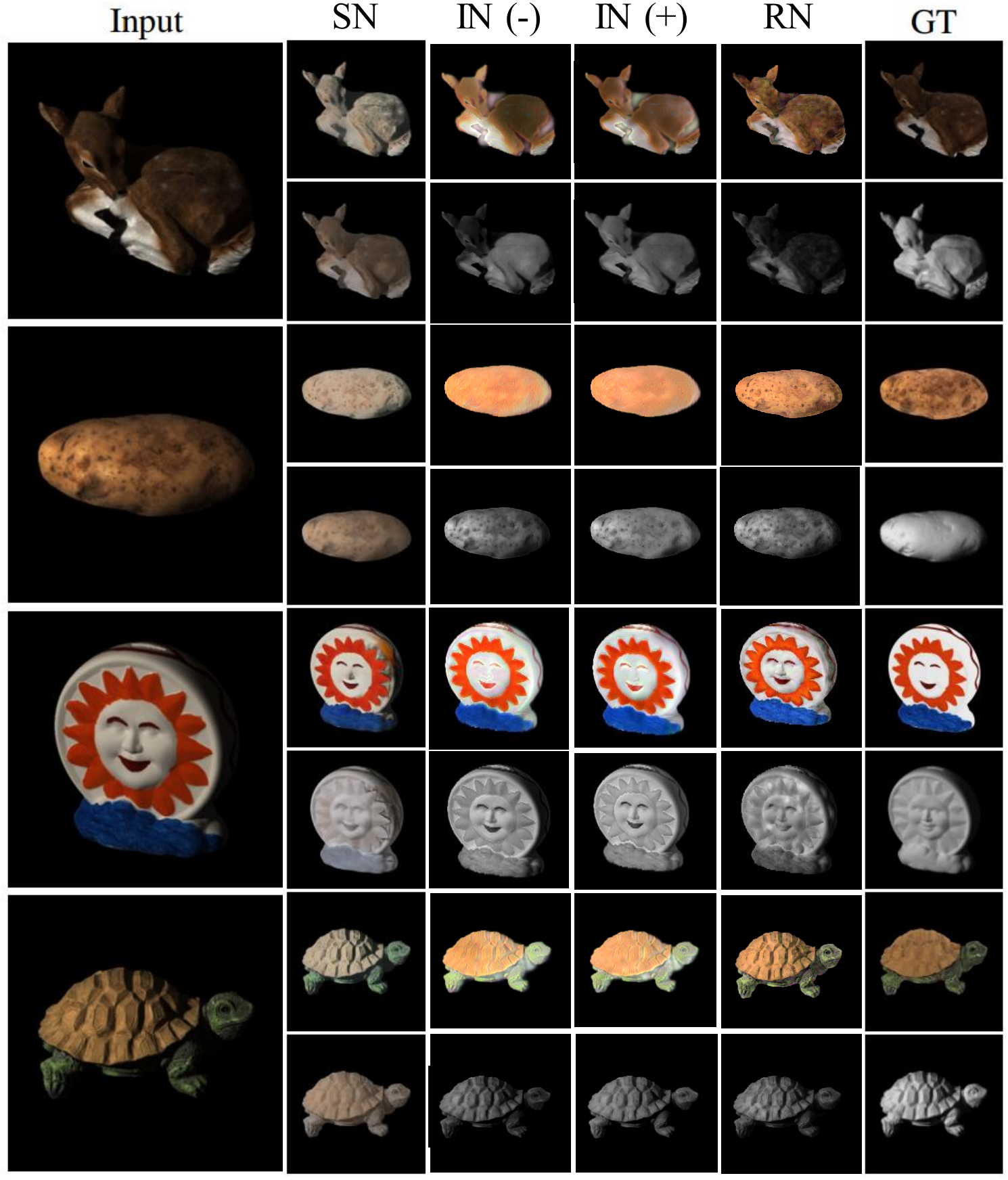}
\centering
\caption{MIT intrinsic benchmark differentiated by the different models. SN is the ShapeNet model of \cite{shapenet}, IN(+/-) denotes the IntrinsicNet with/without the image formation loss, and RN denotes the RetiNet model (including the image formation loss). Proposed models properly recover the reflectance and shading information. IntrinsicNet without the image  formation loss generates color artifacts, and both IntrinsicNets create blurry results compared  with  RetiNet.}
\label{fig:p4c}
\end{figure}

\begin{figure}[t]
\includegraphics[scale=0.66]{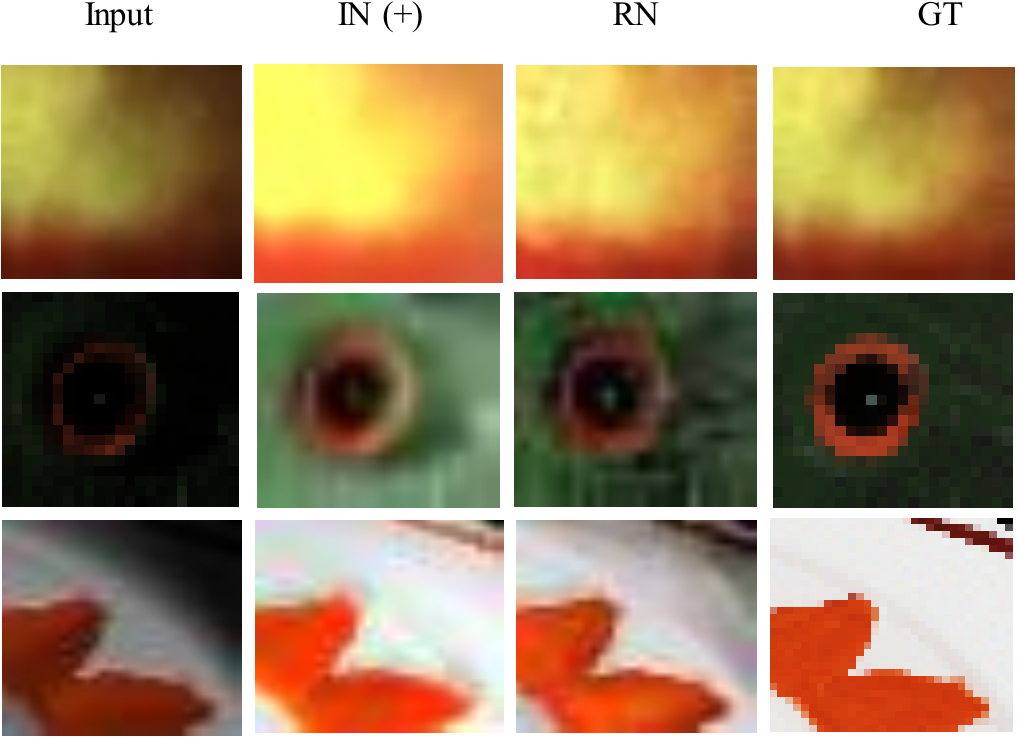} 
\centering
\caption{MIT intrinsic benchmark differentiated by the different models. IN(+) is the IntrinsicNet with the image formation loss, and RN denotes the RetiNet model (including the image formation loss). In RetiNet colors appear more vivid in the reflectance image and it suppresses most of the remaining color artifacts and  blurriness  that  are  present  in  IntrinsicNets.}
\label{fig:p4322}
\end{figure}

To assess our model on real world images, the MIT intrinsic image dataset \cite{mit} is used. The dataset consists of 20 object-centered images with a single canonical light source. Figure \ref{fig:p4c} displays (visual) results and Table \ref{tab:MIT} shows the numeric comparison to other state-of-the-art approaches. Our proposed methods yield better results compared with the ShapeNet~\cite{shapenet} and DirectIntrinsics~\cite{narihia1} models. It can be derived that our proposed models properly recover the reflectance and shading information. However, IntrinsicNet without the image formation loss generates color artifacts, and both IntrinsicNets create blurry results compared with RetiNet. In addition, if an image contains a strong shadow cast, as in the \textit{deer} image, models struggle to eliminate it from the reflectance image. On the other hand, in RetiNet colors appear more vivid in the reflectance image and it suppresses most of the remaining color artifacts and blurriness that are present in IntrinsicNets. Figure \ref{fig:p4322} displays a detailed analysis of RetiNet. 

\begin{table}[t]
  \centering
  \scalebox{0.65}{
  \begin{tabular}{ c||c|c||c|c||c|c }
    \multirow{2}{*}{ } &
      \multicolumn{2}{c||}{MSE}  &
      \multicolumn{2}{c||}{LMSE} &
      \multicolumn{2}{c}{DSSIM} \\
       & Albedo & Shading & Albedo & Shading & Albedo & Shading \\ \hline
    Retinex\cite{mit} & \textbf{0.0032} & 0.0348 & 0.0353 & 0.1027 & 0.1825 & 0.3987 \\ \hline \hline
    DirectIntrinsics\cite{narihia1} & 0.0277 & 0.0154 & 0.0585 & 0.0295 & 0.1526 & 0.1328 \\
    ShapeNet\cite{shapenet} & 0.0468 & 0.0194 & 0.0752 & 0.0318 & 0.1825 & 0.1667 \\ \hline \hline
    IntrinsicNet & 0.0051 & \textbf{0.0029} & \textbf{0.0295} & \textbf{0.0157} & 0.0926 & \textbf{0.0441}\\
    RetiNet & 0.0128 & 0.0107 & 0.0652 & 0.0746 & 0.0909 & 0.1054 \\  \hline \hline
    RetiNet + GT$\nabla$  & 0.0072 & 0.0034 & 0.0429 & 0.0224 & \textbf{0.0550} & 0.0443 \\
  \end{tabular}
  }
  \newline
  \caption {Evaluation results on MIT intrinsic benchmark. Our proposed methods yield better results compared with other models. Experiment with intrinsic gradient ground-truths shows the benefits of exploiting them.}
  \label{tab:MIT}
\end{table}

\subsection{Real and In-the-wild Images}
\begin{figure}[t]
\begin{center}
\includegraphics[width=.9\linewidth]{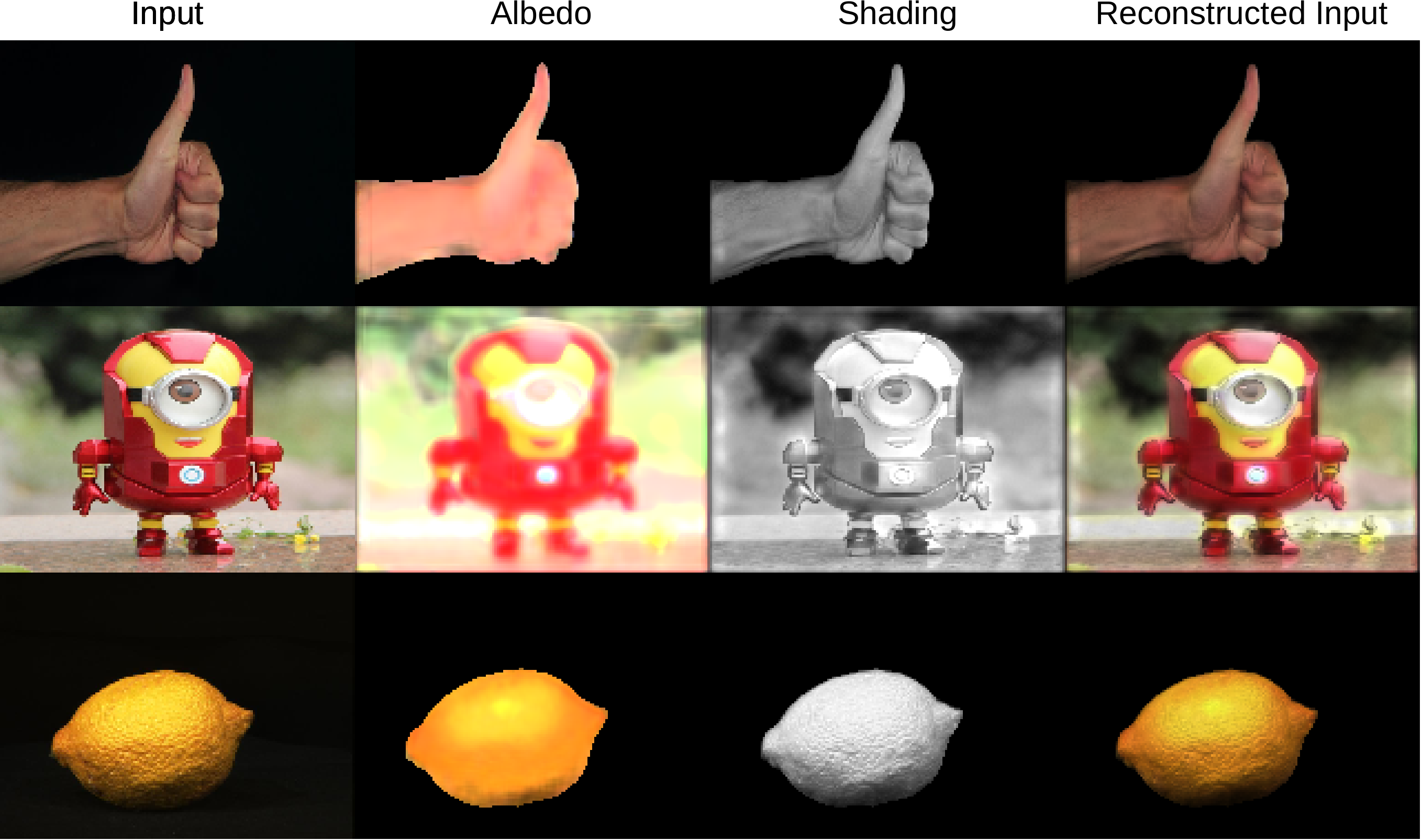}
\end{center}
   \caption{RetiNet applied on real images. It can capture proper albedo image, free of shadings due to geometry.}
\label{fig:real}
\end{figure}

We also evaluate our RetiNet algorithm on real and in-the-wild images. Figure \ref{fig:real} shows the performance of our method (RetiNet) for a number of images. The results show that it can capture proper reflectance image, free of shadings caused by geometry. Finally, we present the reconstructed input from its albedo and shading prediction to show that the decomposition is consistent.


\section{Conclusion}

We proposed two deep learning models considering a physics-based reflection model and gradient information to steer the learning process. The contributions of the paper are as follows. 1: New is the physics-based image formation model in the design of the loss functions. 2: A novel, end-to-end solution is proposed to the well-known Retinex approach based on derivatives. 3: New is the gradient separation part of the RetiNet model in which albedo and shading gradients are learned using a CNN. 4: A (re)integration part is introduced where images are integrated based on gradients by a set of simple convolutions. To train the models, an object centered large-scale synthetic dataset with intrinsic ground-truth images was created. Proposed models were evaluated on synthetic, real world and in-the-wild images. The evaluation results demonstrated that the new model outperforms existing methods. Furthermore, visual inspection showed that the image formation loss function augments color reproduction and the use of gradient information produces sharper edges. Future work will include all intrinsic components in the learning model.

{\scriptsize \noindent \textbf{Acknowledgements:} This project was funded by the EU Horizon 2020 program No. 688007 (TrimBot2020). We thank Partha Das for his contribution to the experiments, Gjorgji Strezoski for his contributions to the figures and the website, and Berkay Kicanaoglu, Mert Kilickaya and Sezer Karaoglu for the fruitful discussions.}

{\small
\bibliographystyle{ieee}
\bibliography{egbib}

\begin{thebibliography}{10}\itemsep=-1pt

\bibitem{Barron0}
J.~T. Barron and J.~Malik.
\newblock Intrinsic scene properties from a single rgb-d image.
\newblock In {\em IEEE Conferance on Computer Vision and Pattern Recognition},
  2013.

\bibitem{Barrow}
H.~G. Barrow and J.~M. Tenenbaum.
\newblock Recovering intrinsic scene characteristics from images.
\newblock {\em Computer Vision Systems}, pages 3--26, 1978.

\bibitem{Beigpour}
S.~Beigpour and J.~van~de Weijer.
\newblock Object recoloring based on intrinsic image decomposition.
\newblock In {\em IEEE International Conference on Computer Vision}, 2011.

\bibitem{bell}
M.~Bell and W.~T. Freeman.
\newblock Learning local evidence for shading and reflectance.
\newblock In {\em IEEE International Conference on Computer Vision}, 2001.

\bibitem{iiw}
S.~Bell, K.~Bala, and N.~Snavely.
\newblock Intrinsic images in the wild.
\newblock {\em ACM Trans. on Graphics (TOG)}, 2014.

\bibitem{Bousseau}
A.~Bousseau, S.~Paris, and F.~Durand.
\newblock User-assisted intrinsic images.
\newblock {\em ACM Trans. on Graphics (SIGGRAPH Asia)}, 2009.

\bibitem{sintel}
D.~J. Butler, J.~Wulff, G.~B. Stanley, and M.~J. Black.
\newblock A naturalistic open source movie for optical flow evaluation.
\newblock In {\em European Conference on Computer Vision}, 2012.

\bibitem{shapenet_dataset}
A.~X. Chang, T.~Funkhouser, L.~Guibas, P.~Hanrahan, Q.~Huang, Z.~Li,
  S.~Savarese, M.~Savva, S.~Song, H.~Su, J.~Xiao, L.~Yi, and F.~Yu.
\newblock Shapenet: An information-rich 3d model repository.
\newblock {\em arXiv preprint arXiv:1512.03012}, 2015.

\bibitem{Chen}
Q.~Chen and V.~Koltun.
\newblock A simple model for intrinsic image decomposition with depth cues.
\newblock In {\em IEEE International Conference on Computer Vision}, 2013.

\bibitem{Duchene}
S.~Duch\^ene, C.~Riant, G.~Chaurasia, J.~L. Moreno, P.~Y. Laffond, S.~Popov,
  A.~Bousseau, and G.~Drettakis.
\newblock Multi-view intrinsic images of outdoors scenes with an application to
  relighting.
\newblock {\em ACM Trans. on Graphics (TOG)}, 2015.

\bibitem{genericmodel}
Q.~Fan, J.~Yang, G.~Hua, B.~Chen, and D.~Wipf.
\newblock A generic deep architecture for single image reflection removal and
  image smoothing.
\newblock In {\em IEEE International Conference on Computer Vision}, 2017.

\bibitem{Funt}
B.~V. Funt, M.~S. Drew, and M.~Brockington.
\newblock Recovering shading from color images.
\newblock In {\em European Conference on Computer Vision}, 1992.

\bibitem{Gehler}
P.~V. Gehler, C.~Rother, M.~Kiefel, L.~Zhang, and B.~Schölkopf.
\newblock Recovering intrinsic images with a global sparsity prior on
  reflectance.
\newblock In {\em Advances in Neural Information Processing Systems}, 2011.

\bibitem{rcnn}
R.~Girshick, J.~Donahue, T.~Darrell, and J.~Malik.
\newblock Rich feature hierarchies for accurate object detection and semantic
  segmentation.
\newblock In {\em IEEE Conferance on Computer Vision and Pattern Recognition},
  2014.

\bibitem{mit}
R.~Grosse, M.~K. Johnson, E.~H. Adelson, and W.~T. Freeman.
\newblock Ground truth dataset and baseline evaluations for intrinsic image
  algorithms.
\newblock In {\em IEEE International Conference on Computer Vision}, 2009.

\bibitem{msra}
K.~He, X.~Zhang, S.~Ren, and J.~Sun.
\newblock Delving deep into rectifiers: Surpassing human-level performance on
  imagenet classification.
\newblock In {\em IEEE International Conference on Computer Vision}, 2015.

\bibitem{batchnorm}
S.~Ioffe and C.~Szegedy.
\newblock Batch normalization: Accelerating deep network training by reducing
  internal covariate shift.
\newblock In {\em International Conference on Machine Learning}, 2015.

\bibitem{kim}
S.~Kim, K.~Park, K.~Sohn, and S.~Lin.
\newblock Unified depth prediction and intrinsic image decomposition from a
  single image via joint convolutional neural fields.
\newblock In {\em IEEE Conferance on Computer Vision and Pattern Recognition},
  2016.

\bibitem{Laffont1}
P.~Y. Laffont and J.~C. Bazin.
\newblock Intrinsic decomposition of image sequences from local temporal
  variations.
\newblock In {\em IEEE International Conference on Computer Vision}, 2015.

\bibitem{Land}
E.~H. Land and J.~J. McCann.
\newblock Lightness and retinex theory.
\newblock {\em Journal of Optical Society of America}, pages 1--11, 1971.

\bibitem{Lee}
K.~J. Lee, Q.~Zhao, X.~Tong, M.~Gong, S.~Izadi, S.~U. Lee, P.~Tan, and S.~Lin.
\newblock Estimation of intrinsic image sequences from image+depth video.
\newblock In {\em European Conference on Computer Vision}, 2012.

\bibitem{coco}
T.~Y. Lin, M.~Maire, S.~Belongie, J.~Hays, P.~Perona, D.~Ramanan,
  P.~Doll{\'a}r, and C.~L. Zitnick.
\newblock Microsoft coco: Common objects in context.
\newblock In {\em European Conference on Computer Vision}, 2014.

\bibitem{skipconnection}
X.~Mao, C.~Shen, and Y.~Yang.
\newblock Image restoration using very deep fully convolutional encoder-decoder
  networks with symmetric skip connections.
\newblock In {\em Advances in Neural Information Processing Systems}, 2016.

\bibitem{Matsushita}
Y.~Matsushita, S.~Lin, S.~B. Kang, and H.~Y. Shum.
\newblock Estimating intrinsic images from image sequences with biased
  illumination.
\newblock In {\em European Conference on Computer Vision}, 2004.

\bibitem{mayer}
N.~Mayer, E.~Ilg, P.~H{\"a}usser, P.~Fischer, D.~Cremers, A.~Dosovitskiy, and
  T.~Brox.
\newblock A large dataset to train convolutional networks for disparity,
  optical flow, and scene flow estimation.
\newblock In {\em IEEE Conferance on Computer Vision and Pattern Recognition},
  2016.

\bibitem{Meka0}
A.~Meka, M.~Zollhöfer, C.~Richardt, and C.~Theobalt.
\newblock Live intrinsic video.
\newblock {\em ACM Trans. on Graphics (SIGGRAPH)}, 2016.

\bibitem{narihia1}
T.~Narihira, M.~Maire, and S.~X. Yu.
\newblock Direct intrinsics: Learning albedo-shading decomposition by
  convolutional regression.
\newblock In {\em IEEE International Conference on Computer Vision}, 2015.

\bibitem{narihia0}
T.~Narihira, M.~Maire, and S.~X. Yu.
\newblock Learning lightness from human judgement on relative reflectance.
\newblock In {\em IEEE Conferance on Computer Vision and Pattern Recognition},
  2015.

\bibitem{synthia}
G.~Ros, L.~Sellart, J.~Materzynska, D.~Vazquez, and A.~M. Lopez.
\newblock The synthia dataset: A large collection of synthetic images for
  semantic segmentation of urban scenes.
\newblock In {\em IEEE Conferance on Computer Vision and Pattern Recognition},
  2016.

\bibitem{imagenet}
O.~Russakovsky, J.~Deng, H.~Su, J.~Krause, S.~Satheesh, S.~Ma, Z.~Huang,
  A.~Karpathy, A.~Khosla, M.~Bernstein, A.~C. Berg, and L.~Fei-Fei.
\newblock Imagenet large scale visual recognition challenge.
\newblock {\em International Journal of Computer Vision}, pages 211--252, 2015.

\bibitem{Shafer}
S.~Shafer.
\newblock Using color to separate reflection components.
\newblock {\em Color research and applications}, pages 210--218, 1985.

\bibitem{Shen}
L.~Shen, P.~Tan, and S.~Lin.
\newblock Intrinsic image decomposition with non-local texture cues.
\newblock In {\em IEEE Conferance on Computer Vision and Pattern Recognition},
  2008.

\bibitem{Shen3}
L.~Shen, X.~Yang, X.~Li, and Y.~Jia.
\newblock Intrinsic image decomposition using optimization and user scribbles.
\newblock {\em IEEE Trans. on Cybernetics}, pages 425--436, 2013.

\bibitem{Shen2}
L.~Shen and C.~Yeo.
\newblock Intrinsic images decomposition using a local and global sparse
  representation of reflectance.
\newblock In {\em IEEE Conferance on Computer Vision and Pattern Recognition},
  2011.

\bibitem{shapenet}
J.~Shi, Y.~Dong, H.~Su, and S.~X. Yu.
\newblock Learning non-lambertian object intrinsics across shapenet categories.
\newblock In {\em IEEE Conferance on Computer Vision and Pattern Recognition},
  2017.

\bibitem{vggnet}
K.~Simonyan and A.~Zisserman.
\newblock Very deep convolutional networks for large-scale image recognition.
\newblock In {\em International Conference on Learning Representations}, 2015.

\bibitem{tappenn}
M.~F. Tappen, E.~H. Adelson, and W.~T. Freeman.
\newblock Estimating intrinsic component images using non-linear regression.
\newblock In {\em IEEE Conferance on Computer Vision and Pattern Recognition},
  2006.

\bibitem{tappen}
M.~F. Tappen, W.~T. Freeman, and E.~H. Adelson.
\newblock Recovering intrinsic images from a single image.
\newblock In {\em Advances in Neural Information Processing Systems}, 2003.

\bibitem{demon}
B.~Ummenhofer, H.~Zhou, J.~Uhrig, N.~Mayer, E.~Ilg, A.~Dosovitskiy, and
  T.~Brox.
\newblock Demon: Depth and motion network for learning monocular stereo.
\newblock In {\em IEEE Conferance on Computer Vision and Pattern Recognition},
  2017.

\bibitem{Weiss}
Y.~Weiss.
\newblock Deriving intrinsic images from image sequences.
\newblock In {\em IEEE International Conference on Computer Vision}, 2001.

\bibitem{Yan}
X.~Yan, J.~Shen, Y.~He, and X.~Mao.
\newblock Retexturing by intrinsic video.
\newblock In {\em Proc. DICTA}, 2010.

\bibitem{Ye}
G.~Ye, E.~Garces, Y.~Liu, Q.~Dai, and D.~Gutierrez.
\newblock Intrinsic video and applications.
\newblock {\em ACM Trans. on Graphics (SIGGRAPH)}, 2014.

\bibitem{Zhao}
Q.~Zhao, P.~Tan, Q.~Dai, L.~Shen, E.~Wu, and S.~Lin.
\newblock A closed-form solution to retinex with nonlocal texture constraints.
\newblock {\em IEEE Trans. on Pattern Analysis and Machine Intelligence}, pages
  1437--1444, 2012.

\bibitem{zhou}
T.~Zhou, P.~Kr\"ahenb\"uhl, and A.~A. Efros.
\newblock Learning data-driven reflectance priors for intrinsic image
  decomposition.
\newblock In {\em IEEE International Conference on Computer Vision}, 2015.

\bibitem{zoran}
D.~Zoran, P.~Isola, D.~Krishnan, and W.~T. Freeman.
\newblock Learning ordinal relationships for mid-level vision.
\newblock In {\em IEEE International Conference on Computer Vision}, 2015.

\end{thebibliography}
}

\end{document}